\documentclass[10pt,twocolumn,letterpaper]{article}

 \usepackage[pagenumbers]{iccv} 

\definecolor{iccvblue}{rgb}{0.21,0.49,0.74}
\usepackage[pagebackref,breaklinks,colorlinks,allcolors=iccvblue]{hyperref}

\usepackage{url}
\usepackage{booktabs}       
\usepackage{amsfonts}       
\usepackage{nicefrac}       
\usepackage{microtype}      
\usepackage{xcolor}         

\usepackage{amsmath}
\usepackage{graphicx}
\usepackage{enumitem}
\usepackage{makecell}
\usepackage{multirow}
\usepackage{arydshln}
\usepackage{subcaption}
\usepackage{amssymb}
\usepackage{pifont}
\usepackage[accsupp]{axessibility}  

\newcommand{\cmark}{\ding{51}}%
\newcommand{\xmark}{\ding{55}}%
\usepackage{color,soul}

\usepackage{array}
\newcolumntype{$}{>{\global\let\currentrowstyle\relax}}
\newcolumntype{^}{>{\currentrowstyle}}
\newcommand{\rowstyle}[1]{\gdef\currentrowstyle{#1}%
  #1\ignorespaces
}

\def\paperID{5933} 
\def\confName{ICCV}
\def\confYear{2025}

\title{Long-LRM: Long-sequence Large Reconstruction Model for Wide-coverage Gaussian Splats}

\vspace{-0.2in}
\author{
Chen Ziwen$^1$\thanks{Research partially done when Chen Ziwen was an intern at Adobe Research}~~~
Hao Tan$^2$~~~ 
Kai Zhang$^2$~~~
Sai Bi$^2$~~~
Fujun Luan$^2$~~~
Yicong Hong$^2$~~~\\
Li Fuxin$^1$~~~
Zexiang Xu$^3$~~~\\
\vspace{-0.1in}
\normalsize
$^1$Oregon State University~~~
$^2$Adobe Research~~~
$^3$Hillbot\\
}

\newcommand{\methodname}{Long-LRM}
\newcommand{\projectpage}{\url{http://arthurhero.github.io/projects/llrm/}}

\begin{document}
\maketitle

\begin{abstract}
We propose \methodname{}, a feed-forward 3D Gaussian reconstruction model for instant, high-resolution, 360$^\circ$ wide-coverage, scene-level reconstruction. Specifically, it takes in 32 input images at a resolution of $960\!\times\! 540$ and produces the Gaussian reconstruction in just 1 second on a single A100 GPU.  
To handle the long sequence of \textbf{250K} tokens brought by the large input size,
\methodname{} features a mixture of the recent Mamba2 blocks and the classical transformer blocks, enhanced by a light-weight token merging module and Gaussian pruning steps that balance between quality and efficiency.
We evaluate \methodname{} on the large-scale DL3DV benchmark and Tanks\&Temples, demonstrating reconstruction quality comparable to the optimization-based methods while achieving an \textbf{800}$\times$ speedup w.r.t. the optimization-based approaches and an input size at least $\mathbf{60} \times$ larger than the previous feed-forward approaches.
We conduct extensive ablation studies on our model design choices for both rendering quality and computation efficiency.
We also explore \methodname{}'s compatibility with other Gaussian variants such as 2D GS, which enhances \methodname{}'s ability in geometry reconstruction. 
Project page: \projectpage
\end{abstract}

\vspace{-0.2in}

\begin{figure}[h]
\vskip -0.15in
    \hspace{-0.15in}
    \includegraphics[width=1.05\linewidth]{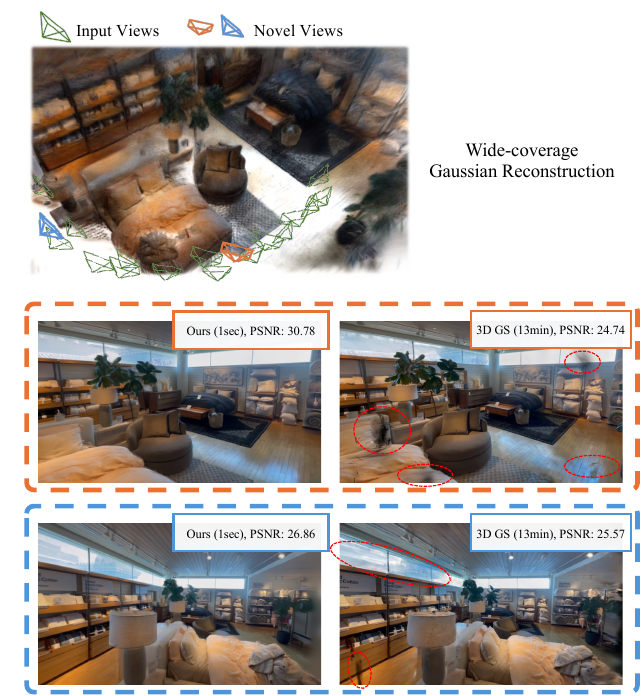}
    \caption{\small \methodname{} reconstructs a real-world scene with wide viewing coverage from 32 images at a resolution of $960\times 540$ in just 1 second. 
    As a feed-forward generalizable model, \methodname{} achieves instant Gaussian reconstruction with high rendering quality comparable to the optimization-based 3D Gaussian splatting methods, which require over 10 minutes to converge on each scene.
    }
    \label{fig:teaser}
    \vspace{-0.2in}
\end{figure}

\section{Introduction}

3D reconstruction from multi-view images is a fundamental problem in computer vision, with applications ranging from 3D content creation, VR/AR, to autonomous driving and robotics. 
Recently, NeRF \citep{nerf} and various radiance field-based methods \citep{instantngp,pointnerf,chen2022tensorf,zipnerf} have shown great potential in reconstructing high-quality 3D scenes from a set of posed images via differentiable rendering. However, these models are slow to reconstruct and not generalizable to unseen scenes, as they require optimization for each scene independently. 
While 3D Gaussian splatting (GS) \citep{3dgs} has significantly advanced reconstruction and rendering efficiency, it still typically requires at least 10 minutes to optimize for each scene and cannot achieve instant reconstruction. 

To address this, generalizable 3D GS models \citep{szymanowicz2023splatter,tang2024lgm} have been proposed to enable fast feed-forward GS reconstruction, avoiding per-scene optimization. Several methods \citep{pixelsplat,gslrm,mvsgaussian,mvsplat} have shown promising scene-level reconstruction results on real 3D captures by regressing per-pixel Gaussian primitives.
In particular, GS-LRM \citep{gslrm}, following the principles of 3D large reconstruction models (LRMs) \citep{hong2024lrm,li2023instant3d,wang2023pflrm} and using transformer blocks \citep{transformer}
without 3D inductive biases such as epipolar attention or sweeping volumes, 
has achieved state-of-the-art novel-view rendering quality on multiple challenging datasets. 
However, 
these previous feed-forward GS models are designed to handle only a small number of input images (typically 1-4) with limited viewing coverage, 
and thus are incapable of reconstructing large real-world scenes, which require at least dozens of images to capture a wide view span. In such cases, optimization-based methods were still the only viable option.

To this end, we propose \methodname{}, which achieves \emph{high-resolution, 360$^\circ$ wide-coverage scene-level Gaussian reconstruction from 32 input images at 960$\times$540 resolution in just 1 second} on a single A100 80G GPU. The 
photorealistic novel-view renderings produced by our approach have a quality comparable to or even better than 3D GS \citep{3dgs} that takes over 10 minutes for per-scene optimization (Fig.~\ref{fig:teaser}).

As inspired by GS-LRM, we patchify the multi-view input images into a sequence of tokens and consider the task of GS reconstruction as a sequence-to-sequence translation to regress pixel-aligned Gaussian primitives. However, unlike GS-LRM that focuses on 2-4 input images, our input setting with 32 960$\times$540 images corresponds to an extremely long token sequence -- \textbf{about 250K tokens} (considering a patch size of $8\!\times\!8$) -- even longer than many modern large language models, such as LLama3~\citep{llama} with a context length of 128K.


This long token length is highly challenging for transformer blocks due to their quadratic time complexity. To address this challenge, we leverage recent advancements in state space models (SSMs) \citep{mamba}, designed to handle long-context reasoning efficiently with linear complexity. 
In particular, we propose a novel LRM architecture that combines Mamba2 \citep{mamba2} blocks with transformer blocks, enabling efficient sequential long-context reasoning while preserving critical global context. 
Additionally, we introduce a light-weight token merging module to further reduce the number of tokens in the middle of the network processing, along with a Gaussian pruning step to encourage efficient use of the dense per-pixel Gaussians. 
These combined designs allow us to train our \methodname{} using similar computational resources to GS-LRM, while successfully scaling up the token length by more than $60\times$, enabling fast, high-quality, wide-coverage reconstruction of real-world scenes.

We train our \methodname{} on the recent DL3DV10K dataset \citep{dl3dv}, which comprises approximately 10K diverse indoor and outdoor scenes. We evaluate our model on both the DL3DV140 benchmark and the Tanks and Temples dataset \citep{tanks}, using 32 input images for each scene. Our direct feed-forward reconstruction achieves comparable novel view synthesis quality to the optimization-based 3D GS approaches, while substantially reducing the reconstruction time (1 second vs. 13 minutes). We conduct extensive ablation studies on our model design choices for both rendering quality and computation efficiency. We also explore \methodname{}'s compatibility with other GS variants such as 2D GS~\cite{2dgs}. \textbf{\methodname{} is the first feed-forward GS solution for wide-coverage scene-level reconstruction in a few seconds.}

\vspace{-0.2in}


\section{Related Work}

\textbf{3D Reconstruction.} Many traditional 3D reconstruction methods have focused on pure geometry reconstruction, where surface meshes~\citep{atlas,neuralrecon,transformerfusion,vortx} or depth maps~\citep{cnnstereo,colmap,mvsnet,cheng2020deep,mvsmachine,deepvideomvs,simplerecon} are the target output. These methods usually involve explicit feature matching along the epipolar lines, followed by TSDF or depth prediction performed by neural networks. In contrast, we leverage the 3D GS representation to simultaneously reconstruct geometry and color, enabling photo-realistic novel view synthesis.

\textbf{Neural reconstruction and rendering.} 
Instead of directly predicting surface geometry, NeRF~\citep{nerf} proposes to leverage differentiable volume rendering to regress novel view images, supervised with a rendering loss.
This implicit way of reconstruction eliminates the need for hard-to-obtain ground-truth 3D supervision while producing visually pleasing reconstruction results. 
However, NeRF requires optimizing an MLP network for each scene independently, which takes hours or even days.
Follow-up works have introduced advanced neural scene representations \citep{zipnerf,instantngp, chen2022tensorf, pointnerf, nerfacto, mip360}, significantly improving time and memory efficiency. Among these, 3D Gaussian splatting~\citep{3dgs} stands out for reducing reconstruction time to just dozens of minutes while maintaining high reconstruction quality and enabling real-time rendering. Variants of 3D GS~\citep{citygaussian,octreegs,vastgaussian} further extend its capabilities to large-scale city-level reconstruction, 
but they still require minutes to optimize on each scene.



\textbf{Generalizable NeRF and 3D GS.} 
Previous attempts to develop generalizable NeRF models have primarily relied on classical projective geometric structures, such as epipolar lines~\citep{pixelnerf,ibrnet,neuray,gprn} or plane-sweep cost volumes~\citep{mvsnerf,geonerf,enerf,zhang2022nerfusion}, to aggregate multi-view features from nearby views for local NeRF estimation. 
Recently, similar designs have been adapted to enable feed-forward scene-level 3D GS reconstruction with generalizable models~\citep{pixelsplat,mvsplat,mvsgaussian}. 
However, since both epipolar geometry and plane-sweep volumes depend on significant overlap between input views, these GS-based methods (as well as most prior NeRF-based methods) are limited to local reconstructions from a small number (1-4) of narrow-span inputs.
On the other hand, GS-LRM \citep{gslrm} avoids these 3D-specific structural designs and adopts global-attention transformer blocks, achieving state-of-the-art performance in this domain. 
However, constrained by the quadratic complexity of transformers, GS-LRM focuses on local reconstruction from only 2-4 views. 
On the other hand, Gamba~\citep{gamba} and MVGamba~\citep{mvgamba} utilize Mamba-based architectures with linear time complexity for object-level GS reconstruction. But they cannot perform scene-level reconstruction and still focus on only 1-4 input views. Our model is instead a novel hybrid model that combines transformer and Mamba2 blocks, designed for 360$^\circ$ wide-coverage, scene-level reconstruction from 32 high-resolution images.

\textbf{Efficient models for long sequences.} 
Transformer-based 3D large reconstruction models (LRMs) \citep{hong2024lrm,li2023instant3d,xu2023dmv3d,wang2023pflrm,wei2024meshlrm,xie2024lrm,gslrm} have emerged for enabling high-quality 3D reconstruction from sparse-view inputs.
While transformers dominate various AI fields due to their flexibility with input modalities and scalability in model sizes, their quadratic time complexity makes them extremely slow when handling long sequences, often requiring thousands of GPUs for parallel computing \citep{llama}.
Efficient architectures such as linear attention~\citep{linear} and structured state space model (SSM)~\citep{s4} are thus proposed in NLP to deal with a large corpus of text.
Mamba~\citep{mamba}, a variant of SSM, offers significant performance improvements by computing state parameters from each input in the sequence and has been successfully extended to tackle vision tasks \citep{visionmamba,vmamba,jamba,localmamba,gamba,mvgamba,hamba}.
Mamba2~\citep{mamba2} further restricts the state matrix $A$ and expands the state dimensions, showing performance comparable to transformers on multiple language tasks.
However, empirical studies~\citep{empirical} indicate that transformers still outperform Mamba2 in in-context learning and long-context reasoning, both critical for 3D reconstruction.
Inspired by \cite{empirical} and Jamba~\citep{jamba}, we propose a hybrid architecture combining transformer and Mamba2 blocks for long token-sequence 3D GS reconstruction, achieving a balance between training efficiency and reconstruction quality.


\section{Method}
\label{sec:method}

\begin{figure*}
    \centering
    \includegraphics[width=1.0\linewidth]{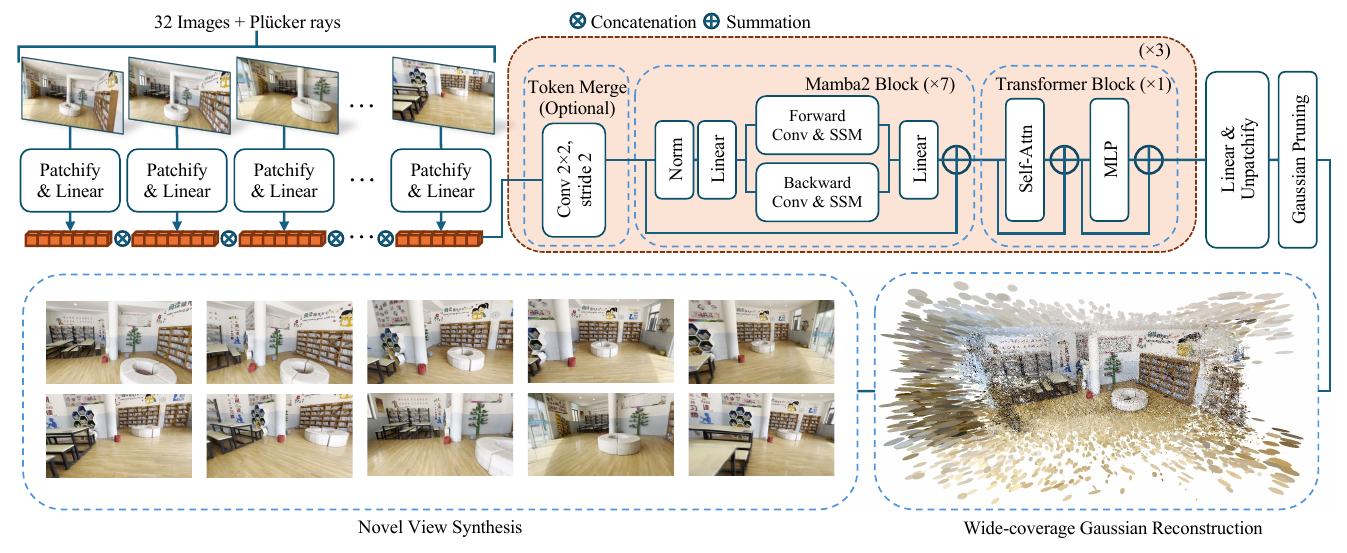}
    \vspace{-0.3in}
    \caption{\small
    \methodname{} takes 32 input images along with their Plücker ray embeddings as model input, which are then patchified into a token sequence. These tokens are processed through a series of  Mamba2 and transformer blocks (\{7M1T\}$\times$3). 
    Fully processed, the tokens are decoded into per-pixel Gaussian parameters, followed by a Gaussian pruning step. The bottom section illustrates the resulting wide-coverage Gaussian reconstruction and photo-realistic novel view synthesis.}
    \label{fig:method}
    \vspace{-0.15in}
\end{figure*}

In this section, we present our \methodname{} method.
We give an overview in Sec.~\ref{sec:method_overall_arch}, the implementation details of the Mamba2 blocks in Sec.~\ref{sec:method_mamba2} and additional efficiency improvement designs (e.g., token merging) in Sec.~\ref{sec:method_token_merge}.
We end with a discussion of the training objectives in Sec.~\ref{sec:method_losses} that help the model to effectively converge.

\subsection{Overall Architecture}
\label{sec:method_overall_arch}

As shown in Fig.~\ref{fig:method}, we 
first tokenize the channel-wise concatenated RGB images and Plücker rays. Similar to GS-LRM~\citep{gslrm}, we view the per-pixel GS prediction as a sequence-to-sequence mapping. 
But crucially, we use a hybrid of Mamba2 blocks and transformer blocks, following the studies in \cite{empirical} and \cite{jamba}, for better scalability to higher resolution and denser views.
In our implementation, each hybrid block consists of 7 Mamba blocks and one transformer block, which we empirically observe to be a balanced configuration and easier to converge. 
For the transformer blocks, we use global self-attention. 
We detail our implementation of Mamba2 blocks in Sec.~\ref{sec:method_mamba2}.
A token merging module is injected between the hybrid blocks to further speed up the processing (Sec.~\ref{sec:method_token_merge}).


We decode per-pixel Gaussian parameters from the output tokens using a linear layer and apply training-time and test-time Gaussian pruning to improve efficiency at high resolution settings.



\subsection{Mamba2 Block}
\label{sec:method_mamba2}
A Mamba block~\citep{mamba}, similar to a transformer block, processes a token sequence of shape $L\!\times\!D$ by mixing the token information, and outputs a token sequence of the same shape.
For a sequence of length $L$, the transformer block has a computational complexity of $O(L^2)$ while Mamba effectively reduces it to $O(L)$.
Thus, it is suitable for the dense reconstruction task in our \methodname{}.

As a variant of SSM, Mamba processes each input token $x_t$ by formula $h_{t}=\textbf{A}h_{t-1}+\textbf{B}x_t$ and $y_t=\textbf{C}h_{t}$,
where $h_t$ is the hidden state, $y_t$ is the output token, $t$ is the sequence index, and $\textbf{A},\textbf{B},\textbf{C}$ are parameters. Different from previous work~\citep{s4}, Mamba computes $\textbf{A},\textbf{B},\textbf{C}$ from the input 
instead of storing them as model parameters. 
The novel Mamba2~\citep{mamba2} block improves over Mamba by restricting the state matrix $\mathbf{A}$ to be a scalar times an identity matrix, allowing the usage of efficient block multiplication and expansion to larger state dimensionality, showing performance comparable to transformers on multiple language tasks.
However, since the Mamba2 block is designed for language tasks, it only scans through the tokens in one direction, which is suboptimal for images. Following \citep{visionmamba}, we take bi-directional scans over the concatenated token sequence. Specifically, we first compute the state parameters from the input using one linear layer; then we run the SSM block in both forward and backward directions on the token sequence. Finally, we sum up the output tokens from the two scans before going through another linear layer.
We have conducted some preliminary exploration of more complex scan patterns as in VMamba~\citep{vmamba} and LocalMamba~\citep{localmamba}, but decided not to adopt those due to a substantial decrease in speed. 


\subsection{Token Merging and Gaussian pruning}
\label{sec:method_token_merge}

With 32 960$\times$540 input images and patch size 8, the sequence length can reach 250K, highly challenging even for linear-complexity models like Mamba. Empirically, we find even the all-Mamba2 variant of our model runs out of memory under the highest resolution setting (Tab.~\ref{tb:archi}). To further reduce memory usage, we propose to merge the tokens in the middle of the network.

Our light-weight token merging module is inspired by the traditional multi-level CNN encoders and effectively reduces token sequence length down to 1/4. We first reshape the token sequence from $L\!\times\!D$ back to $N\!\times\!\frac{H}{p}\!\times\!\frac{W}{p}\!\times\!D$ where $p$ is the patch size. Then, we apply a channel-wise $2\!\times\!2$ 2D convolution with stride 2, resulting in output shape $N\!\times\!\frac{H}{2p}\!\times\!\frac{W}{2p}\!\times\!D'$, where $D'$ is the new token dimensionality that can differ from the original one. Finally, we reshape it back to $\frac{L}{4}\!\times\!D'$ where each token now has an `effective' patch size of $2p$. We conduct ablation studies on the position of token merging in the network (Tab.~\ref{tb:tm}) and the effect of token dimensionality change (Tab.~\ref{tb:archi}), showing a balance between computational efficiency and reconstruction quality.

However, even with token merging, the per-pixel Gaussian prediction still brings an enormous quantity of Gaussians ($\sim$17 million for 32 images with resolution 960$\times$540), posing intense pressure on memory and speed. There is likely a huge redundancy because of the high resolution of inputs and the overlap between the input camera frustums. 
Thus, a Gaussian pruning step is necessary. 

Naively pruning away the most transparent Gaussians , even the ones with only 5\% opacity, results in incomplete \textit{opaque} objects in the reconstructed scene, hinting that the model tends to diffuse an opaque color into multiple translucent Gaussians.
To encourage the model to use a compact set of Gaussians, we apply a punishment on the opacity of all Gaussians (detailed in Sec.~\ref{sec:method_losses}).
With the reduced number of visible Gaussians, we can thus simply prune away Gaussians with low opacity.
Empirically, we find no difference in rendering quality if removing Gaussians with opacity below 0.001.
We apply Gaussian pruning before rendering in both training and inference at the  960$\times$540 resolution setting.
We use a fixed-percentage pruning instead of an opacity threshold to ensure constant training memory usage.

\subsection{Training Objectives}
\label{sec:method_losses}

Below, we list the training objectives for \methodname:

\noindent\textbf{Rendering loss.} Following previous work~\citep{gslrm}, we use a combination of Mean Squared Error (MSE) loss and Perceptual loss
\begin{equation}
\footnotesize
    \mathcal{L}_\text{image}=\frac{1}{M}\sum_{i=1}^M \left( \text{MSE}\left(\mathbf{I}^{\text{gt}}_i, \mathbf{I}^{\text{pred}}_i\right) + \lambda \cdot \text{Perceptual}\left(\mathbf{I}^{\text{gt}}_i, \mathbf{I}^{\text{pred}}_i\right)  \right)
\end{equation}
to supervise the quality of the rendered images, where $\lambda$ is set to 0.5. While training solely with rendering loss can already achieve competitive visual quality (see Sec.~\ref{sec:ablation}), we further introduce two regularization terms to improve training stability and computational efficiency.

\noindent\textbf{Depth regularization for training stability.}
Training instability is a well-known curse for large-scale training.
In our task, we observe that the instability comes from the difficulty in optimizing the Gaussian positions.
With rendering loss only, the model will produce ill-posed Gaussians known as ``floaters", which do not lie on the actual 3D surface -- a common issue for novel view synthesis (see the black ``floaters" in Fig.~\ref{fig:teaser}).
To stabilize training, we add a scale-invariant depth loss
\begin{equation}
\small
    \mathcal{L}_\text{depth}=\frac{1}{M}\sum_{i=1}^M \text{Smooth-L1}\left(\mathbf{D}^{\text{da}}_i, \mathbf{D}^{\text{pred}}_i\right) 
\end{equation}
where $\mathbf{D}^{\text{da}}_i$ is the disparity map predicted by DepthAnything~\citep{depthanything}, and $\mathbf{D}^{\text{pred}}_i$ is the disparity map obtained from the predicted position of the per-pixel Gaussians. Following \cite{depthanything}, we normalize the disparity maps by subtracting their medians $t(d_i)$ and then dividing by their mean absolution deviation $\frac{1}{HW}\sum|d_i-t(d_i)|$. This soft depth supervision effectively helps reduce the chance of the training divergence.

\noindent\textbf{Opacity regularization for inference efficiency.}
Since our per-pixel prediction strategy renders a dense set of Gaussians, to encourage an efficient use of the Gaussians, we apply a small L1 regularization on the opacity
\begin{equation}
\small
     \mathcal{L}_\text{opacity}= \frac{1}{N}\sum_{i=1}^N |o_i|
\end{equation}
where the opacity values are between 0 and 1. 
Intuitively, L1 can encourage the sparsity of the regularized terms~\citep{tibshirani1996regression}.
We empirically observe that adding this loss can drastically push the percentage of Gaussians with opacity above 0.001 from 99\% down to around 40\% (see Tab.~\ref{tb:opa}).
With these near-zero opacity Gaussians, we can perform Gaussian pruning as discussed in Sec.~\ref{sec:method_token_merge} and reduce both the Gaussian rendering time and the backpropagation time.


\noindent\textbf{Overall training loss.} Our total loss is thus the rendering loss and the weighted regularization loss terms:
\begin{equation}
\small
    \mathcal{L} =  \mathcal{L}_\text{image}+ \lambda_\text{opacity}\cdot \mathcal{L}_\text{opacity}+ \lambda_\text{depth}\cdot \mathcal{L}_\text{depth}
\end{equation}
where we set $\lambda_\text{opacity}=0.1$ and $\lambda_\text{depth}=0.01$.

\section{Experiments}

\begin{figure*}[h]
    \centering
    \includegraphics[width=\linewidth]{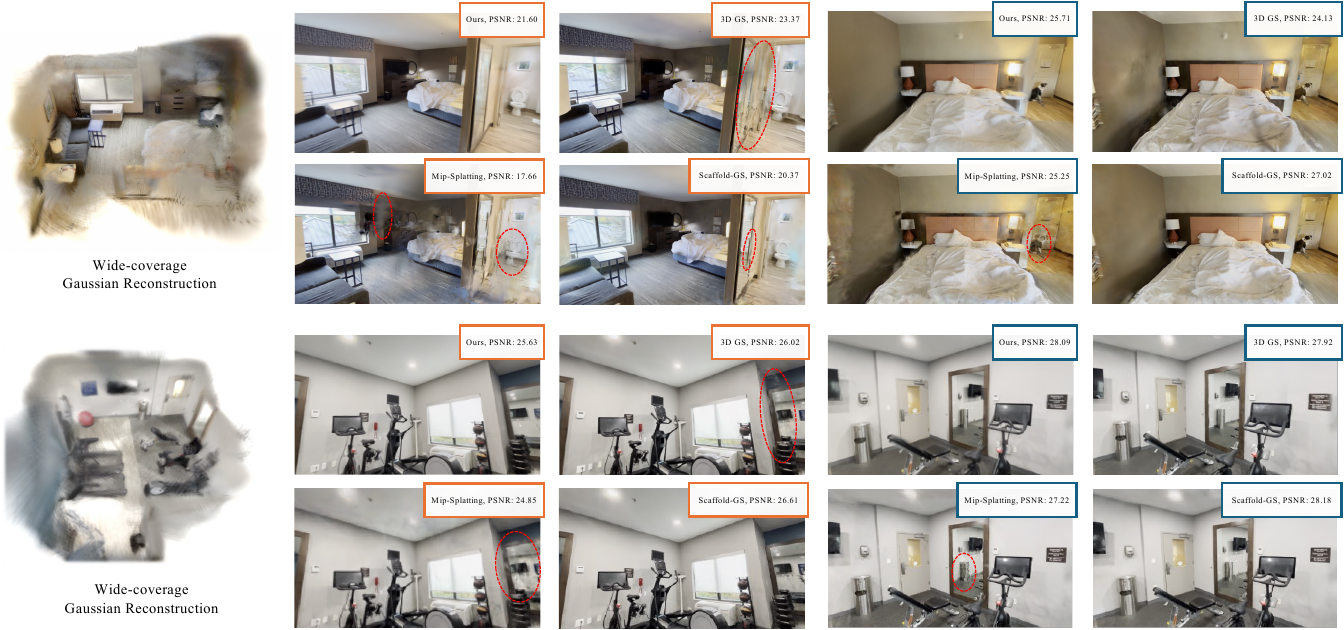}
    \vskip -0.05in
    \caption{\small Qualitative comparisons between \methodname{} (1 second) and optimization-based 3D GS methods (13+ minutes), reconstructed from 32 images at $960\times540$ resolution. The leftmost column shows the overlook of our wide-coverage Gaussian reconstruction, while the rest columns show the color renderings from the reconstructed 3D Gaussians. Our approach maintains high-quality reconstruction with competitive PSNR, demonstrating the ability to generate accurate details and fewer artifacts in challenging regions. Red ellipses highlight areas where optimization-based methods struggle with artifacts or inaccuracies, whereas \methodname{} produces cleaner outputs.}
    \label{fig:qual_comp}
    \vskip -0.15in
\end{figure*}

\subsection{Implementation and Experiment Details}

\paragraph{Architecture Details.} 
Our model consists of 24 blocks, with every 7 Mamba2 blocks followed by 1 transformer block, repeating 3 times. We start with a patch size of 8 and a token dimensionality of 256. We perform token merging at the beginning of the 9th block, with patch size expanded to 16 and token dimensionality expanded to 1024.  For Mamba2 blocks, we use a 256-dimensional state, an expansion rate of 2 and a head dimensionality of 64. For transformer blocks, we use a head dimensionality of 64 and an MLP dimension ratio of 4. We use the FlashAttentionV2~\citep{flashattv2} implementation that optimizes GPU IO utilization for long sequences.
\paragraph{Training Settings.}
Directly training the model on high-resolution images is extremely inefficient; therefore, we opt for an low-to-high-resolution curriculum training schedule, with three training stages, using image resolutions of $256\!\times\!256$, $512\!\times\!512$ and $960\!\times\! 540$.

Specifically, in the first stage, training images are resized so that the shorter side is 256 and then center-cropped to a square shape. 
For the training view selection, we first randomly pick a consecutive subsequence ranging from 64 frames to 128 frames, then uniformly sample 32 images as input and randomly sample 8 images as target.
Input and target are sampled independently and thus can overlap. We randomly shuffle or reverse the input view order with probability 0.5.
We train with a peak learning rate of $4\textsc{e}{-4}$ and the AdamW optimizer~\citep{adamw} with a weight decay of $0.05$. 
The learning rate is linearly warmed up in the first 2K steps and then cosine decayed.
We use a batch size of 256 and train for 60K steps.

In the 2nd stage, we resize and crop the images to $512\!\times\! 512$, decrease the peak learning rate to $4\textsc{e}{-5}$, and train the model for 10K steps at batch size 64. 
The view selection protocol remains the same.

In the last stage, we resize the images to $960\!\times\! 540$ without square cropping, expand the view sampling range to the entire sequence (about 200$\sim$300 frames for DL3DV), and keep training the model for another 10K steps with batch size 64. 
We perform Gaussian pruning at this stage to save GPU memory usage, where we only keep top 40\% Gaussians ranked by opacity, plus 10\% randomly sampled from the rest.
We augment the FOV of the images by randomly center-cropping the images to 0.77$\sim$1.0 of the original size and resize back, in order to fit a broader range of camera models.

\paragraph{Evaluation Settings.}

During evaluation, our goal is to reconstruct the scene captured by the entire video sequence. 
Following the protocol of previous novel-view synthesis works~\citep{mip360,3dgs}, we uniformly pick every 8-th image of the sequence as the test split. 
From the rest of the sequence, we use $K$-means clustering (based on camera positions and viewing directions) to pick a set of input views that can best cover the entire scene. 
The number of clusters is set to the number of input views. 
The cameras closest to the cluster centers are chosen as the input split. 
We use an image resolution of $960\times 540$ during the evaluation. 
We perform Gaussian pruning during evaluation by only keeping the top 50\% of the Gaussians with highest opacity values, which is an empirically safe range with negligible quality loss. 

\paragraph{Post-prediction Optimization.} We provide a post-prediction optimization option to further improve reconstruction quality. After \methodname{} predicts the initial set of Gaussians, we render at the input camera views, compute MSE loss between the rendered images and the input images, and back-propagate gradients to the Gaussian parameters. We set a learning rate of $5\textsc{e}{-4}$ for position, $1\textsc{e}{-3}$ for color, and keep opacity, scale and rotation unchanged.

\paragraph{Training with 2D GS.} We additionally explore \methodname{}'s compatibility with 2D GS~\cite{2dgs}, a variant of 3D GS known for its strength in geometry reconstruction and depth map rendering. Specifically, after completing the first two training stages, we adapt the original \methodname{} to 2D GS by directly interpreting the predicted Gaussian parameters as 2D Gaussian parameters. We then fine-tune the model for an additional 10K steps at a resolution of $512\!\times\! 512$, followed by the same third-stage training as in the original pipeline.

\subsection{Datasets}

DL3DV~\citep{dl3dv} is a large-scale, real-world scene dataset for 3D reconstruction and novel view synthesis. It features diverse scene types, with both indoor and outdoor captures. The DL3DV-10K split consists of 10,510 high-resolution videos,
each accompanied by 200$\sim$300 keyframes with camera pose annotation obtained from COLMAP~\citep{colmap}; the DL3DV-140 Benchmark split contains 140 test scenes. We train our model on DL3DV-10K and evaluate on the DL3DV-140 Benchmark. We also perform zero-shot inference on Tanks\&Temples~\citep{tanks}, another challenging real-world scene dataset, to show the generalization capability of \methodname{}.
Following previous work~\citep{3dgs,mvsgaussian}, we use the \texttt{train} and the \texttt{truck} scene.
To compare with previous feed-forward GS methods, we also evaluate \methodname{} under a low-resolution, two-view setting on RealEstate10K~\citep{realestate}, a real-world indoor scene dataset, following the same train-test split and evaluation setting introduced by pixelSplat~\cite{pixelsplat}. To test \methodname{}'s instant geometry reconstruction capability after equipped with 2D GS, we perform zero-shot depth map evaluation on ScanNetv2~\cite{scannet} (see supplementary), a classic indoor scene dataset for multi-view stereo.

\subsection{Results}

\begin{figure*}[h]
    \centering
    \includegraphics[width=\linewidth]{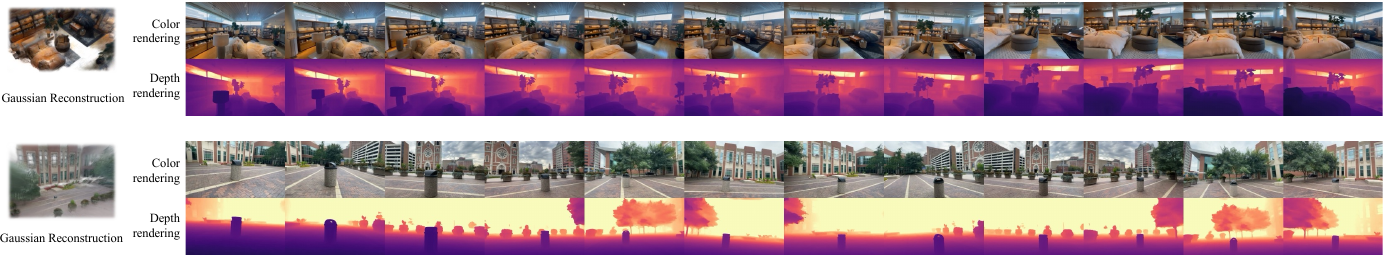}
    \vskip -0.05in
    \caption{\small Qualitative results showcasing \methodname{}'s color and depth reconstruction capabilities on two wide-span scenes (the second being 360$^\circ$). At a sequence of novel camera views, we render RGB images from the 3D Gaussians predicted by \methodname{}, and depth maps from the 2D Gaussians predicted by \methodname{} finetuned with 2D GS. The depth maps exhibit sharp object edges and smooth surfaces, demonstrating \methodname{}'s ability to generalize and inherit strengths from different GS variants.}
    \label{fig:color_depth}
    \vskip -0.15in
\end{figure*}

\begin{table}[h]
\centering

\begin{minipage}[t]{\linewidth}
\begin{center}
    \captionsetup{type=table}
    \hspace{-0.1in}\resizebox{\linewidth}{!}{
\begin{scriptsize}
\renewcommand{\arraystretch}{1.2}
\begin{tabular}{@{}$c@{\ \ }^l@{ }^c@{ }^c@{ }^c@{ }^c@{ }^c@{ }^c@{ }^c@{}}
\toprule
\rowstyle{\bfseries}
\multirow{2}{*}{\makecell{Input\\Views}} & \multirow{2}{*}{Method} &  \multirow{2}{*}{Time$\downarrow$} & \multicolumn{3}{^c}{DL3DV-140} & \multicolumn{3}{^c}{Tanks\&Temples} \\\cmidrule(lr){4-6}\cmidrule(lr){7-9}
 &  & & PSNR$\uparrow$ & SSIM$\uparrow$ & LPIPS$\downarrow$ & PSNR$\uparrow$ & SSIM$\uparrow$ & LPIPS$\downarrow$ \\
 \midrule
\multirow{4}{*}{16} &3D GS$_{30k}$   & 13min & 21.20 & 0.708 & 0.264 & 16.76 & 0.598 & 0.334 \\
&Mip-Splatting$_{30k}$ & 13min & 20.88 & 0.712 & 0.274 & 16.82 & 0.616 & 0.332 \\
&Scaffold-GS$_{30k}$ & 16min & 22.13 & 0.738 & \textbf{0.250} & 17.02 & \textbf{0.634} & \textbf{0.321} \\
&Ours  & \textbf{0.4sec} & \textbf{22.66} & \textbf{0.740} & 0.292 & \textbf{17.51} & 0.555 & 0.408 \\
\midrule
\multirow{6}{*}{32} &3D GS$_{30k}$ & 13min & 23.60 & 0.779 & 0.213 & 18.10 & 0.688 & 0.269 \\
&Mip-Splatting$_{30k}$ &  13min & 23.32 & 0.784 & 0.217 & 18.39 & \textbf{0.700} & \textbf{0.262} \\
&Scaffold-GS$_{30k}$ & 16min & 24.77 & 0.805 & \textbf{0.205} & 18.41 & 0.691 & 0.290 \\
& Ours  & \textbf{1sec} & 24.10 & 0.783 & 0.254 & 18.38 & 0.601 & 0.363 \\
& Ours$_3$  & 12sec & 24.99 & 0.809 & 0.243 &  18.69 & 0.623 & 0.360  \\
& Ours$_{10}$  & 37sec & \textbf{25.60} & \textbf{0.826} & 0.233 & \textbf{18.90} & 0.642 & 0.350 \\
\bottomrule
\end{tabular}
\end{scriptsize}}
\end{center}
\vspace{-0.6cm}
\caption{\footnotesize \textbf{Quantitative comparison to optimization-based 3D GS approaches on full-scene reconstruction.} Subscripts indicate number of optimization steps. Our post-prediction optimization step (3.6sec) optimizes over all input images.
`Time' refers to the total inference/optimization time for each scene. 
Image resolution is $960\times 540$.
}
\label{tb:3dgs}
\end{minipage}
\end{table}

\begin{table}[h]
\centering

\begin{minipage}[t]{0.48\linewidth}
\begin{center}
    \captionsetup{type=table}
    \hspace{-0.1in}\resizebox{\linewidth}{!}{
\begin{scriptsize}
\renewcommand{\arraystretch}{1.2}
\begin{tabular}{@{}$l@{ }^c@{ }^c@{ }^c@{}}
\toprule
\rowstyle{\bfseries}
Method & PSNR$\uparrow$ & SSIM$\uparrow$ & LPIPS$\downarrow$ \\
 \midrule
 pixelSplat & 25.89 & 0.858 & 0.142\\
 MVSplat & 26.39 & 0.869 & 0.128 \\
 GS-LRM & 28.10 & 0.892 & 0.114 \\
Ours  & \textbf{28.54} & \textbf{0.895} & \textbf{0.109} \\
 \bottomrule
\end{tabular}
\end{scriptsize}}
\end{center}
\vspace{-0.6cm}
\caption{\footnotesize Quantitative comparison on RealEstate10K under 2-view setting. Image resolution is $256\times 256$.}
\label{tb:sparse}
\end{minipage}
\hfill
\begin{minipage}[t]{0.5\linewidth}
\begin{center}
    \captionsetup{type=table}
    \hspace{-0.1in}\resizebox{\linewidth}{!}{
\begin{scriptsize}
\renewcommand{\arraystretch}{1.2}
\begin{tabular}{@{}$c@{\ \ }^l@{ }^c@{ }^c@{ }^c@{ }^c@{}}
\toprule
\rowstyle{\bfseries}
\makecell{Input\\Views} & Method &  Time$\downarrow$ & PSNR$\uparrow$ & SSIM$\uparrow$ & LPIPS$\downarrow$  \\
 \midrule
\multirow{3}{*}{32} & 2D GS$_{30k}$ &  13min & 24.35 & \textbf{0.813} & \textbf{0.203} \\
& Ours  & \textbf{1sec} & 23.24 & 0.755 & 0.287 \\
& Ours$_{10}$  & 37sec & \textbf{24.74} & 0.800 & 0.263 \\
\bottomrule
\end{tabular}
\end{scriptsize}}
\end{center}
\vspace{-0.6cm}
\caption{\footnotesize Comparison between \methodname{} (2D GS ver.) and  optimization-based 2D GS~\cite{2dgs} on color reconstruction quality on DL3DV-140 at resolution $960\!\times\! 540$.
}
\label{tb:2dgs}
\end{minipage}
\end{table}

\methodname{} achieves wide-coverage, scene-level 3D Gaussian splatting reconstruction from 32 high-resolution input images, which, to the best of our knowledge, no other method have accomplished. Recent work pixelSplat~\citep{pixelsplat}, MVSplat~\citep{mvsplat}, MVSGaussian~\citep{mvsgaussian}, and GS-LRM~\citep{gslrm} are limited to processing 1–4 input images, with pixelSplat and MVSplat showing results only at $256\!\times\!256$ resolution. Most of these methods rely on traditional 3D inductive biases, such as epipolar projection and cost volumes, which are suited for narrow-view inputs with large overlaps but struggle with wide-coverage, high-resolution settings. Moreover, naively extending these methods to handle more input views and higher resolutions leads to out-of-memory issues and requires significant architectural changes. Therefore, we compare \methodname{} with the \textit{optimization-based 3D GS methods in the high-resolution, wide-coverage setting}, and compare with \textit{previous feed-forward methods in a low-resolution, two-input setting}.

\textbf{High-resolution, wide-coverage reconstruction.} In Table~\ref{tb:3dgs}, we show a quantitative comparison with the optimization-based 3D GS methods \cite{3dgs,mipsplat,scaffoldgs} on two real-world wide-span scene datasets: DL3DV-140~\citep{dl3dv} and Tanks\& Temples~\citep{tanks}. 
We show results under a sparser 16 input-view setting and the 32 input-view setting.
Our feed-forward model achieves an $800\times$ speedup for reconstructing an unseen novel scene (1 second vs. 13 minutes), while our reconstruction quality is comparable with the 30K-step optimization results, and can be further boosted with just a few post-prediction optimization steps. Our model takes the lead in terms of PSNR (+2dB with Ours$_{10}$ vs. 3D GS$_{30k}$), while optimization-based methods perform better in LPIPS. 
Qualitative results (Fig.~\ref{fig:teaser} and~\ref{fig:qual_comp}) show that \methodname{} significantly reduces floater artifacts, which can be attributed to the prior knowledge distilled from a large training dataset and our regularization terms: the opacity loss and the soft depth supervision. More visualization and interactive results can be found on our project page and in the supplementary.

\textbf{Low-resolution, sparse-view reconstruction.}
In Table~\ref{tb:sparse}, we present a quantitative comparison with the feed-forward GS methods \cite{pixelsplat,mvsplat,gslrm} on RealEstate10K at a $256\!\times\!256$ resolution with 2 input views, a setting commonly used in prior works. \methodname{} achieves the best overall quality, outperforming \cite{pixelsplat} and \cite{mvsplat} by a margin of over 2dB PSNR and slightly surpassing the transformer-based GS-LRM, highlighting the effectiveness of our hybrid architecture.

\textbf{Compatibility with other GS variants.} In Table~\ref{tb:2dgs}, we show that after being finetuned for 2D GS prediction, the color rendering quality of \methodname{} is still comparable with the optimization-based counterpart, while Fig.~\ref{fig:color_depth} shows the rendered depth maps from the predicted 2D Gaussians, highlighting \methodname{}'s ability to generalize and inherit strengths from different GS variants. We also perform a zero-shot depth map and mesh quality evaluation on the ScanNetv2 dataset to quantitatively assess the geometry reconstruction ability of \methodname{} (see supplementary).

Overall, \methodname{} not only leads to state-of-the-art rendering quality in the classical sparse-view setting but also enables instant, wide-coverage, high-resolution, scene-level reconstruction that no previous methods can achieve.

\section{Analysis}
\subsection{Ablation Studies of Model Designs}
\label{sec:ablation}

\begin{table*}[h]
\centering
\begin{minipage}[t]{0.72\linewidth}
\begin{center}
    \captionsetup{type=table}
    \hspace{-0.1in}\resizebox{\linewidth}{!}{%
\begin{scriptsize}
\renewcommand{\arraystretch}{1.2}
\begin{tabular}{@{}$c@{ }^c@{ }^c@{ }^c^l@{ }@{ }^c@{ }^c@{ }^c^c@{ }^c@{ }^c@{ }^c@{}}
\toprule
\rowstyle{\bfseries}
\makecell{Input\\Views} & \makecell{Image\\Size} & \makecell{Batch Size\\ / GPU} & \makecell{Train\\Step} & Block Type & \makecell{Token\\Merge} & \makecell{Patch\\Size} & \makecell{Token\\Dimensionality} & \#Param & \makecell{Iteration\\ Time (sec)} & \makecell{GPU \\Memory (GB)} & PSNR$\uparrow$\\
\midrule
\multirow{4}{*}{4} & \multirow{4}{*}{256} & \multirow{4}{*}{16} & \multirow{4}{*}{100K}& Transformer & / & 8 & 1024 & 307M & 2.3 & 44 & 21.13 \\ 
&& && Mamba2 & / & 8 & 1024 & 170M & 2.8 & 35 & 19.82 \\ 
&& && \{7M1T\}$\times 3$ & / & 8 & 1024 & 187M & 2.6 & 35 & \textbf{21.58} \\
&& && \{7M1T\}$\times 3$ & @9 & 8 $\rightarrow$16 & 1024 & 198M & 2.0 & 23 & 20.96 \\
&& && \{7M1T\}$\times 3$ & @9 & 8 $\rightarrow$16 & 256 $\rightarrow$1024 & \textbf{142M} & \textbf{1.9} & \textbf{20} & 20.43 \\
\midrule

\multirow{4}{*}{32} & \multirow{4}{*}{256} & \multirow{4}{*}{4} &\multirow{4}{*}{60K}& Transformer &  / &  8 & 1024 & 307M & 14.5 & 68 &  too slow\\
&&& & Mamba2 &  / &  8 & 1024 & 170M & 6.0 & 70 & 24.28 \\
&&& & \{7M1T\}$\times 3$ &  / &  8 & 1024 & 187M & 7.1 & 70 & \textbf{26.82} \\
&&& & \{7M1T\}$\times 3$ & @9 &  8 $\rightarrow$16 & 1024 & 198M & 5.4 & 35 & 25.81 \\
&&& & \{7M1T\}$\times 3$ & @9 &  8 $\rightarrow$16 & 256 $\rightarrow$1024 & \textbf{142M} & \textbf{3.5} & \textbf{25} & 25.62 \\
\midrule

\multirow{4}{*}{32} & \multirow{4}{*}{512} & \multirow{4}{*}{1} & \multirow{4}{*}{10K$^*$} &  Transformer & / & 8 & 1024 & 307M & 50.5 & 44 & too slow \\
&&&& Mamba2 & / & 8 & 1024 & 170M & 7.4 & 62 & 24.83 \\
&&&& \{7M1T\}$\times 3$ & / & 8 & 1024 & 187M & 11.5 & 64 & \textbf{28.16} \\
&&& & \{7M1T\}$\times 3$ & @9 &  8 $\rightarrow$16 & 1024 & 198M & 7.0 & 60 & 27.69 \\
&&& & \{7M1T\}$\times 3$ & @9 &  8 $\rightarrow$16 & 256 $\rightarrow$1024 & \textbf{142M} & \textbf{4.0} & \textbf{23} & 27.46 \\
\midrule
\multirow{2}{*}{32} & \multirow{2}{*}{$960\!\times\! 540$} & \multirow{2}{*}{1} & \multirow{2}{*}{10K$^*$} &  \multicolumn{8}{^l}{\textbf{All other variants are out of memory.}}  \\
&&& & \{7M1T\}$\times 3$ & @9 &  8 $\rightarrow$16 & 256 $\rightarrow$1024 & 142M & 12.6 & 53 & 27.32 \\
\bottomrule
\end{tabular}
\end{scriptsize}}
\end{center}
\vspace{-0.3cm}
\caption{\small \textbf{Ablation studies on model architecture.} We study how the model architecture affects training time and memory efficiency as well as the reconstruction quality. 
\textbf{\{7M1T\}$\mathbf{\times 3}$} refers to our ``7 Mamba2 blocks + 1 transformer block, repeating 3 times" model architecture. 
\textbf{@9} refers to the placement of token merging in front of the 9th block. 
``$\mathbf{\rightarrow}$" indicates the change accompanying the token merging step.
$^*$The 512-resolution models are finetuned from the checkpoints of their 256-resolution counterparts, and the 960-resolution from the 512-resolution checkpoints.
}
\label{tb:archi}
\end{minipage}\hfill
\begin{minipage}[t]{0.26\linewidth}
\begin{center}
    \captionsetup{type=table}
    \hspace{-0.1in}\resizebox{\linewidth}{!}{%
\begin{scriptsize}
\renewcommand{\arraystretch}{1.2}
\begin{tabular}{@{}$c@{ }^c@{ }^c@{ }^c@{ }^c@{}}
\toprule
\rowstyle{\bfseries}
\makecell{Token\\Merge} & Time (sec) & Mem (GB) & \#Param & PSNR$\uparrow$ \\
\midrule
@1 & 2.0 & 21G & 198M & \textbf{21.35} \\
@9 & \textbf{1.9} & 20G & 142M & 21.25 \\
@17 & \textbf{1.9} & \textbf{19G} & \textbf{86M} & 20.99 \\
\bottomrule
\end{tabular}
\end{scriptsize}}
\end{center}
\vspace{-0.6cm}
\caption{\footnotesize \textbf{Ablation studies on placement of token merging in the network} with 4-view 256-resolution setup.
}
\label{tb:tm}
\vspace{-0.3cm}
\begin{center}
    \captionsetup{type=table}
    \hspace{-0.1in}\resizebox{\linewidth}{!}{%
\begin{scriptsize}
\renewcommand{\arraystretch}{1.2}
\begin{tabular}{@{}$l@{ }^c@{ }^c@{}}
\toprule
\rowstyle{\bfseries}
Loss Type & PSNR$\uparrow$ & \makecell{\% Gaussians w/\\ opacity$>$0.001}\\
\midrule
rendering-only &  20.43 & 99.2\\\
 +opacity & 20.96 & 68.3 \\ 
 +opacity+depth & 21.25 & 70.1\\ 
\bottomrule
\end{tabular}
\end{scriptsize}}
\end{center}
\vspace{-0.6cm}
\caption{\footnotesize \textbf{Ablation studies on training objectives} with 4-view 256-resolution setup. We study how the opacity loss and the depth supervision affect the reconstruction quality and the Gaussian usage.
}
\label{tb:loss}
\vspace{-0.3cm}
\begin{center}
    \captionsetup{type=table}
    \hspace{-0.1in}\resizebox{\linewidth}{!}{%
\begin{scriptsize}
\renewcommand{\arraystretch}{1.2}
\begin{tabular}{@{}$c@{ }^c@{ }^c@{ }^c@{ }^c@{}}
\toprule
\rowstyle{\bfseries}
\makecell{Input\\Views} & \makecell{Image\\Size} & \makecell{Input Sampling\\ Range (frame)}  &  \makecell{w/ opacity\\loss} & \makecell{\% Gaussians w/\\ opacity$>$0.001}\\
\midrule
4 & $256\!\times\! 256$ &  16 & \xmark& 99.2 \\
4 & $256\!\times\! 256$ & 16 & \cmark& 68.3 \\
\hdashline
32 & $256\!\times\! 256$ & $64\sim 128$ & \cmark& 41.8 \\ 
32 & $512\!\times\! 512$  & $64\sim 128$ & \cmark& 34.1 \\
32 & $960\!\times\! 540$ & \makecell{$200\sim 300$} & \cmark& 33.3 \\
\bottomrule
\end{tabular}
\end{scriptsize}}
\end{center}
\vspace{-0.6cm}
\caption{\footnotesize \textbf{Gaussian usage impacted by opacity loss and input size}.
}
\label{tb:opa}

\end{minipage}
\vskip -0.05in
\end{table*}

In Table~\ref{tb:archi}, we show how the model architecture variants scale with the input size in terms of both training efficiency and reconstruction quality.
We consider 4 experimental setups: 1. sparse low-resolution (`Input Views'=4, `Image Size'=256), 2. dense low-resolution (`Input Views'=32, `Image Size'=256), 3. dense high-resolution (`Input Views'=32, `Image Size'=512), 4. dense ultra-resolution (`Input Views'=32, `Image Size'=960$\times$540) ~\footnote{Note that here the terminology of `sparse', `dense', `low', `high', `ultra' are all relative. We use these terminology for simplicity and clarity.}.
The results of different model architecture under the same setup are presented within a Table block.
We evaluate five model variants: (1) all Transformer blocks (equivalent to GS-LRM), (2) all Mamba2 blocks, (3) a hybrid model without token merging, (4) a hybrid model with token merging and a constant token dimensionality of 1024, and (5) our final model—a hybrid approach with token merging, starting with a token dimensionality of 256. All variants consist of 24 blocks in total.
For each variant, we report the number of parameters (\#Param), training iteration time, GPU memory usage, and PSNR reconstruction performance. The detailed experimental setup is provided in the supplementary.

\noindent\textbf{All transformer blocks.}
The performance of transformer is comparable to the hybrid model under the 4-view 256-resolution setting.
However, its training time explodes for larger inputs, either with denser views or higher resolutions.
Under the 32-view 512-resolution setup, the per-iteration time with batch size=1 reaches the unaffordable 50.5 seconds due to the quadratic time complexity.


\noindent\textbf{All Mamba2 blocks.}
The Mamba2 variant shows a more manageable increase in time but leads to a noticeable decline in reconstruction quality.
In the 256-resolution, 4-view setup, the Mamba2 variant exhibits a 1.8 PSNR drop compared to our hybrid model, and the gap widens with longer sequences, reaching 2.5 PSNR for 32 views and 3.3 PSNR at 512 resolution, which is possibly due to Mamba's state-based design, which struggles to capture long-range dependencies.

\noindent\textbf{Impact of Token Merging.}
Our hybrid model gets the best of both worlds -- the reconstruction quality of transformer and the speed of Mamba2. However, a token merging step is essential for training in the most challenging dense ultra-resolution setup. Reducing the token length to 1/4 while keeping a constant token dimensionality of 1024 reduces training time to 2/3 in the dense-view setup. Further reducing the initial token dimensionality to 256—scaling up to 1024 only after merging—saves an additional 1/3 of training time and memory, making ultra-resolution training feasible, where all other variants run out of memory. We analyze the effect of token merging placement in the network in Tab.~\ref{tb:tm}, aiming at a balance between model size and reconstruction quality.






\subsection{Ablation Studies of Training Objectives}
\label{sec:ablation}

\noindent\textbf{Impact of the regularization terms.}
Table~\ref{tb:loss} presents the impact of the two regularization terms introduced in Sec.~\ref{sec:method_losses}: opacity loss and depth supervision. We find that adding opacity loss significantly reduces the number of visible Gaussians (measured as the \% of Gaussians with opacity above 0.001) while having a negligible effect on rendering performance.
Depth supervision further improves rendering quality and helps prevent gradient explosions by guiding ``floater" Gaussians toward the true surface. This encourages the model to use SH coefficients to represent color variations based on the viewing direction, rather than relying on translucent floater Gaussians that act as lens filters in front of the cameras.

\noindent\textbf{Gaussian Usage.}
Table~\ref{tb:opa} presents the impact of the opacity loss and input size on Gaussian usage (measured as the \% of Gaussians with opacity above 0.001). 
As the input resolution increases—leading to more per-pixel Gaussians being predicted—the chance that multiple pixels can be covered by one Gaussian increases as well, causing a decrease in Gaussian usage.
Meanwhile, when the viewing span increases and the overlap between input views decreases, the model needs to retain more Gaussians to maintain reconstruction quality. This counteracts the effect of higher resolution, resulting in only a negligible drop in Gaussian usage in the last row.



\section{Conclusions}
In this work, we present \methodname{}, a novel approach for fast and scalable 3D Gaussian reconstruction.  By combining Mamba2 and transformer blocks, along with token merging and Gaussian pruning, \methodname{} achieves instant wide-span scene reconstruction from 32 images at a high resolution of $960\times540$ in just 1 second, delivering rendering quality comparable to optimization-based 3D GS methods. Additionally, \methodname{} outperforms previous feed-forward GS methods in low-resolution, sparse-view setups and demonstrates compatibility with other GS variants, such as 2D GS. Through extensive ablation studies, we highlight the advantages of our hybrid architecture, the efficiency gains from token merging, and the effectiveness of the proposed regularization terms.

\section*{Acknowledgements}

We thank Nathan Carr and Kalyan Sunkavalli for their support and helpful discussions. We thank Lu Ling for the support in using the DL3DV dataset.
Chen Ziwen and Li Fuxin are partially supported by Oregon State University seed grant AGD010-AS06 and NSF 2321851.

{
    \small
    \bibliographystyle{ieeenat_fullname}
    \bibliography{main}
}

\clearpage

\appendix

\begin{figure*}[h]
    \centering
    \includegraphics[width=\linewidth]{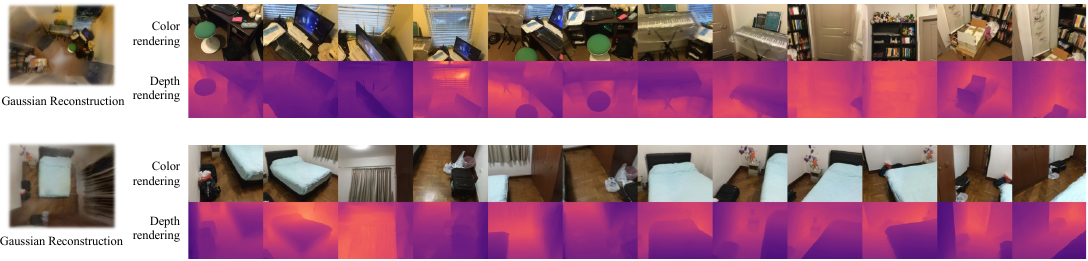}
    \vskip -0.05in
    \caption{\small Zero-shot reconstruction results on ScanNetv2 from \methodname{} (2D GS ver.). We render both RGB images and depth maps from the predicted 2D Gaussians. The depth maps exhibit fine object details, demonstrating \methodname{}'s capability for instant geometry reconstruction on complicated novel scenes.}
    \label{fig:scannet}
\end{figure*}

\begin{figure*}[t]
    \centering
    \includegraphics[width=0.99\linewidth]{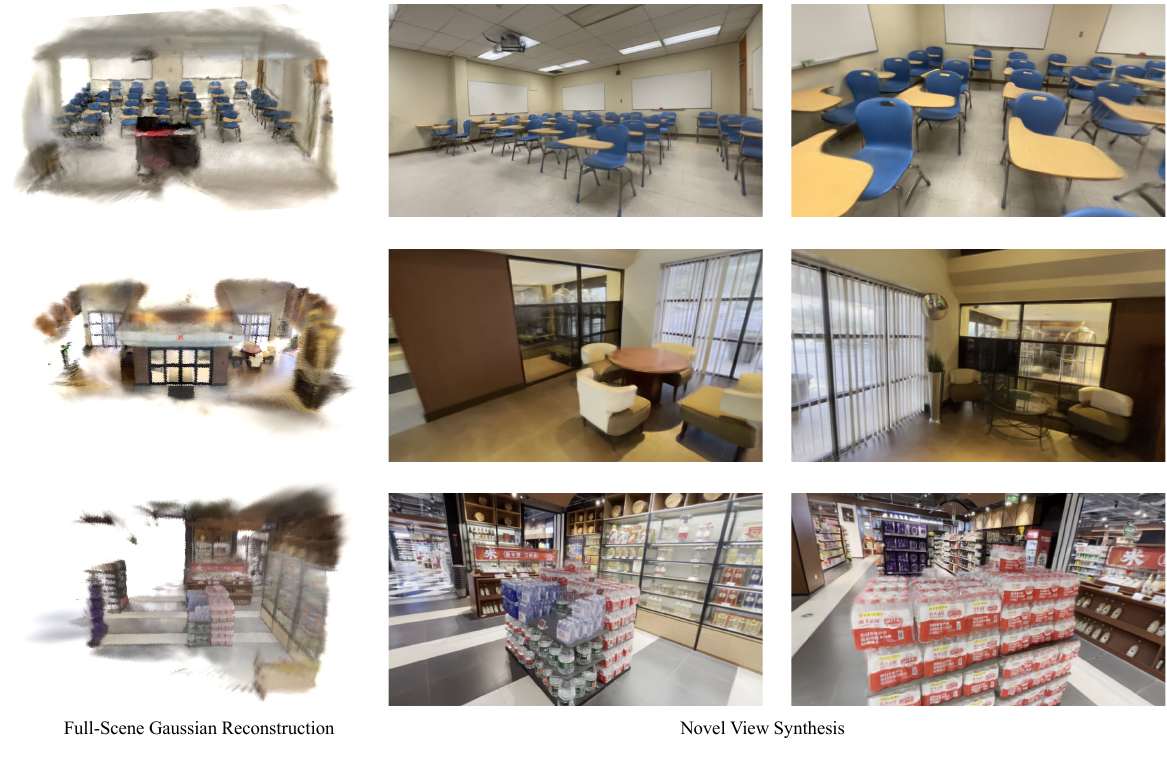}
    \caption{More qualitative results from \methodname{}'s wide-coverage scene reconstruction. The left column illustrates the overlook of the reconstructed Gaussians, and the right columns show high-quality synthesized novel views from different perspectives. These examples demonstrate \methodname{}'s ability to handle diverse and complex scenes, accurately reconstructing fine-level details, and generating photorealistic views from multiple angles, effectively capturing both geometric and appearance variations.}
    \label{fig:results}
\end{figure*}

\section{More Qualitative Results}
We show more qualitative results of \methodname{} in Fig.~\ref{fig:results} and project page (\projectpage), including a video comparison with optimization-based 3D GS methods.


\section{More Hyperparameter Settings}

\textbf{Gaussian parameters.} We use SH degree 3 for predicted Gaussian colors. We apply a bias -6.9 and a maximum cap -1.2 to the Gaussian scales before sending them to the exponential function. We apply a bias -2.0 to the opacity values before sending them to the sigmoid function. We align the per-pixel Gaussians to the camera rays originated from the pixels. 

\noindent\textbf{Camera pose normalization.} In the dataloader, we calculate the average pose of the input cameras by averaging the forward-, downward- and rightward camera directions, and use the cross-product method to obtain an orthonormal rotation matrix along with the average camera positions in the world coordinate. We use the inverse of this average camera pose to normalize all cameras. Finally, we rescale the camera positions to a [-1,1] bounding box.

\section{Experiment Details for Ablation Studies on Model Architecture}

In Table~\ref{tb:archi}, we present the model architecture ablation studies with different length of input sizes.
We train all variants on DL3DV-10K  and evaluate on DL3DV-140.
The number of training steps are empirically decided based on the model convergence.
We study the model behavior under four different settings: 4 input views at $256\!\times\! 256$, 32 input views at $256\!\times\! 256$, and 32 input views at $512\!\times\! 512$, and our extreme setting: 32 input views at $960\!\times\! 540$.

For these ablation studies, we use a shorter frame range during evaluation for fair comparisons among each experiments.
In details, we choose the first 96 frames from the original video frame sequence, then uniformly sample 8 test views.
The 4 to 32 training views are then uniformly sampled from the rest views, i.e., not overlapping to the testing views. 
We keep the same set of training and testing views for different experimental setups.
The input images are resized and center-cropped to squares except for the last row.

\section{Additional Experiment Results}

\begin{figure}
    \centering
    \includegraphics[width=0.99\linewidth]{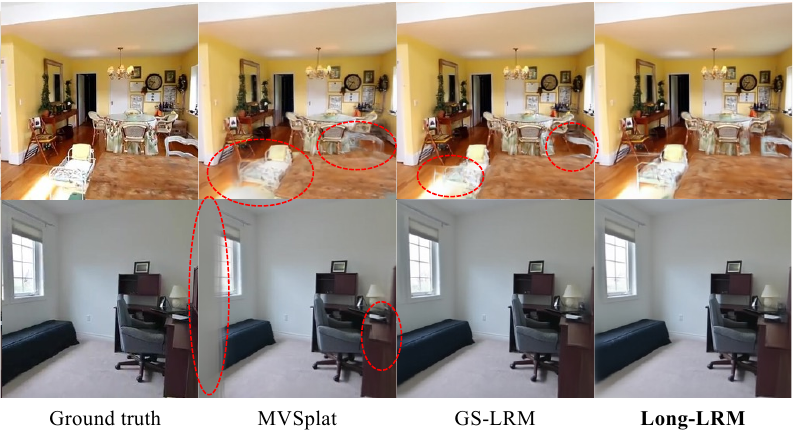}
    \vspace{-0.1in}
    \caption{\footnotesize Qualitative comparison on RealEstate10K.}
    \label{fig:re10k}
\end{figure}

\begin{table}[h]
\centering

\begin{minipage}[t]{0.8\linewidth}
\begin{center}
    \captionsetup{type=table}
    \hspace{-0.1in}\resizebox{\linewidth}{!}{
\begin{scriptsize}
\renewcommand{\arraystretch}{1.2}
\begin{tabular}{@{}$l@{ }^c@{ }^c@{ }^c@{ }^c@{}}
\toprule
\rowstyle{\bfseries}
 Method & Init & PSNR$\uparrow$ & Chamfer$\downarrow$ & F-Score$\uparrow$ \\
 \midrule
 2D GS & COLMAP & 19.25 & 0.193 & 0.272\\
 \methodname(2D) & / & \textbf{23.27} &  \textbf{0.135} & \textbf{0.414} \\
\bottomrule
\end{tabular}
\end{scriptsize}}
\end{center}
\vspace{-0.6cm}
\caption{\footnotesize Novel view synthesis (PSNR) and mesh reconstruction quality (Chamfer and F-Score) comparison with optimization-based 2D GS on ScanNetv2 test split.
}
\label{tb:mesh}
\end{minipage}

\end{table}

\begin{table}[h]
\centering

\begin{minipage}[t]{\linewidth}
\begin{center}
    \captionsetup{type=table}
    \hspace{-0.1in}\resizebox{\linewidth}{!}{
\begin{scriptsize}
\renewcommand{\arraystretch}{1.2}
\begin{tabular}{@{}$c@{ }^c@{ }^c@{ }^l@{ }^c@{ }^c@{ }^c@{ }^c@{}}
\toprule
\rowstyle{\bfseries}
 \makecell{Input\\Views} & \makecell{Depth\\Supervision} & \makecell{Zero-\\shot} & Method & Abs Diff$\downarrow$ & Abs Rel$\downarrow$ & Sq Rel$\downarrow$ & $\mathbf{\delta\!<\!1.25}$$\uparrow$ \\
 \midrule
 \multirow{3}{*}{\makecell{$80\!\sim\!800$\\(every 10th)}} & \multirow{3}{*}{\cmark} &  \multirow{3}{*}{\xmark} & COLMAP~\cite{colmap} & 0.264 & 0.137 & 0.138 & 83.4 \\
 & & &Atlas~\cite{atlas} & 0.123 & 0.065 & 0.045 & 93.6 \\
 & & & VoRTX~\cite{vortx} & 0.092 & 0.058 & 0.036 & 93.8 \\
 \midrule
 32 & \xmark & \cmark & \methodname{} & 0.119 & 0.073 & 0.051 & 94.0\\
\bottomrule
\end{tabular}
\end{scriptsize}}
\end{center}
\vspace{-0.6cm}
\caption{\footnotesize We evaluate \methodname{}(2D)'s ability for zero-shot full-scene geometry reconstruction on the ScanNetv2 test split~\cite{scannet}. With only 32 input images, we render median-depth maps from the reconstructed 2D Gaussians and evaluate against the ground-truth depth at \textbf{all} frames in each scene sequence. To put into context, we also list performance of past MVS approaches under the same evaluation settings that use much denser input views and are trained on ScanNet training split with ground-truth depth.
}
\label{tb:depth}
\end{minipage}
\end{table}

\begin{table}[h]
\centering

\begin{minipage}[t]{0.49\linewidth}
\begin{center}
    \captionsetup{type=table}
    \hspace{-0.1in}\resizebox{\linewidth}{!}{
\begin{scriptsize}
\begin{tabular}{@{}$l@{ }^l@{ }^c@{ }^c@{}}
\toprule
\rowstyle{\bfseries}
Method & Loss Type & PSNR$\uparrow$ & \makecell{\% Gaussians w/\\ opacity$>$0.001}\\
\midrule
GS-LRM & rendering-only & 28.10 & 99.9\\
\methodname{} & rendering-only & \textbf{28.54} & 99.9\\
 \methodname{} & +opacity+depth & 28.44 & 44.7\\ 
\bottomrule
\end{tabular}
\end{scriptsize}}
\end{center}\vspace{-0.6cm}
\caption{\footnotesize Ablation studies on training objectives with 2-view 256-resolution setup on RealEstate10K.
}
\label{tb:re10k_loss}
\end{minipage}
\hfill
\begin{minipage}[t]{0.5\linewidth}
\begin{center}
    \captionsetup{type=table}
    \hspace{-0.1in}\resizebox{\linewidth}{!}{
\begin{scriptsize}
\renewcommand{\arraystretch}{1.2}
\begin{tabular}{@{}$l@{ }^c@{ }^c@{ }^c@{}}
\toprule
\rowstyle{\bfseries}
Layers & PSNR$\uparrow$ & Iter Time (sec) & Memory (GB) \\
 \midrule
 \{1T7M\}$\times$3 & 21.62 & 2.9 & 35\\
  \{7M1T\}$\times$3 & 21.58 & 2.6 & 35 \\
 3T21M & diverged & 3.8 & 35 \\
  21M3T & diverged & 3.8 & 35 \\
 \bottomrule
\end{tabular}
\end{scriptsize}}
\end{center}
\vspace{-0.6cm}
\caption{\footnotesize Ablation study on the layer combination of transformer and Mamba2.}
\label{tb:tmarch}
\end{minipage}

\end{table}

\subsection{More results on RealEstate10K.}

We report the performance of \methodname{} with and without depth and opacity losses on RealEstate10K in Tab.~\ref{tb:re10k_loss}. We observe that the losses are less essential for this two-view setup, as the opacity loss aims to save the number of Gaussians and the depth loss helps stabilize training for our long input setup. We also turn off the token merging module for this setup as saving resources is no longer necessary. We show qualitative comparisons in Fig.~\ref{fig:re10k}, where \methodname{} demonstrates better rendering of details than baselines.

\subsection{Zero-shot results on ScanNetv2.}
We present several zero-shot reconstruction results on the ScanNetv2 test split with our model trained solely on DL3DV-10K. We add aspect ratio augmentation in stage 3 of training, where the input image are randomly cropped between 1:1 to 1.77:1. During inference, we sample 32 input images from each scene of ScanNet and resize the images so that the height is 540.

We report in Table~\ref{tb:mesh} the novel view synthesis and the mesh reconstruction quality comparison between \methodname{}(2D GS ver.) and the optimization-based 2D GS. For novel view synthesis, we evaluate on every 80-th frame, given the denser image sequence of ScanNet.
For mesh reconstruction, we render median-depth maps from the reconstructed 2D Gaussians at every 10th camera and use TSDF fusion (voxel 4cm) to construct the mesh. 
Note that the 32-view setup is relatively sparse for ScanNet scenes, leading to poor performance of optimization-based 2D GS, even with the extra COLMAP initialization, stressing the value of the prior knowledge learned by our model.

Table~\ref{tb:depth} reports quantitative results in terms of depth map quality metrics. 
We render depth maps from the predicted 2D Gaussians at all keyframes in the video sequence at the original resolution of $1272\!\times\!948$. To put the results in context, we list the performance of a few classic multi-view stereo approaches that are trained on the ScanNet train split with ground-truth depth supervision under the same evaluation settings. Fig.~\ref{fig:scannet} shows the qualitative results of the color and depth renderings from the predicted 2D Gaussians.

\subsection{Ablation studies on layer combination.}
\methodname{} uses a hybrid architecture of Mamba2 and transformer blocks. In Table~\ref{tb:tmarch} we study the impact of different block combination configurations on model performance. We use the 4-input settings at image resolution $256\times 256$ without token merging. We found that if the transformer blocks are even distributed among the Mamba2 blocks, then the model training is more stable than transformer blocks being concentrated at the beginning or the end of the model. The PSNR curves during training do not differ much for the configurations in the table, including the ones that diverge at the end.

\section{Limitations}
We now briefly discuss the limitations. While we have successfully scaled the model to support 32 high-resolution views and achieved wide-coverage large-scale GS reconstruction, we observe only marginal performance improvements when further increasing the number of input views. Specifically, increasing the input to 64 views only leads to less than 1 dB PSNR improvement. Notably, 64 high-res images correspond to extremely long sequences, exceeding 500K in context length, which presents a significant challenge for current sequence processing models. Addressing this limitation will require future work to better manage ultra-long sequences.
Additionally, since the entire DL3DV training set contains images with a fixed wide field of view (FOV), we found that our model struggles to generalize on test sets with significant FOV variations (e.g., the MipNeRF360 dataset with a much smaller FOV). We suspect this limitation is due to the use of Mamba2 blocks, as differing FOVs can alter the meaning of tokens at different positions. Developing models that can generalize effectively across varying FOVs may require more diverse datasets with a range of various FOVs, at a scale similar to DL3DV.

\end{document}